\documentclass[11pt]{article}

\usepackage[in]{fullpage}
\usepackage{times}
\usepackage{graphicx}
\usepackage[rflt]{floatflt}
\usepackage{epsfig,subfigure,color,multirow}
\usepackage{amsmath,amssymb,algorithm,algorithmic,theorem,float,bbm,bm,enumerate}
\usepackage{url}
\usepackage{epstopdf}

\newcommand{\beq}{\begin{equation}}
\newcommand{\eeq}{\end{equation}}

\newcommand\R{\mathbb{R}}

\renewcommand{\a}{\mathbf{a}}
\renewcommand{\b}{\mathbf{b}}

\renewcommand{\v}{\mathbf{v}}

\newcommand{\x}{\mathbf{x}}
\newcommand{\y}{\mathbf{y}}

\newcommand{\cX}{{\cal X}}

\newcommand{\bU}{\mathbf{U}}

\newcommand{\myref}[1]{(\ref{#1})}

\DeclareMathOperator{\argmin}{argmin}

\newcounter{exampleI}
{\theorembodyfont{\rmfamily} \theoremstyle{plain} }

\newcounter{exampleII}
{\theorembodyfont{\rmfamily} \theoremstyle{plain} }

\newcounter{exampleIII}
{\theorembodyfont{\rmfamily} \theoremstyle{plain} }

{\theorembodyfont{\rmfamily} }
{\theorembodyfont{\rmfamily} }
{\theorembodyfont{\rmfamily} }
\newtheorem{thm}{Theorem}

\newtheorem{lem}{Lemma}
\newtheorem{asm}{Assumption}

\newcommand{\proof}{\noindent{\itshape Proof:}\hspace*{1em}}

\newcommand{\qed}{\nolinebreak[1]~~~\hspace*{\fill} \rule{5pt}{5pt}\vspace*{\parskip}\vspace*{1ex}}

\newcommand {\commentout}[1] {}

\title{ Randomized Block Coordinate Descent \\
for Online and Stochastic Optimization}
\author{Huahua Wang \\
Dept of Computer Science \& Engg\\
University of Minnesota, Twin Cities\\
huwang@cs.umn.edu
\and
Arindam Banerjee\\
Dept of Computer Science \& Engg\\
University of Minnesota, Twin Cities\\
banerjee@cs.umn.edu
}

\date{}

\begin{document}

\maketitle

\begin{abstract}
Two types of low cost-per-iteration gradient descent methods have been extensively studied in parallel. One is online or stochastic gradient descent ( OGD/SGD), and the other is randomzied coordinate descent (RBCD).  In this paper, we combine the two types of methods together and propose online randomized block coordinate descent (ORBCD). At each iteration, ORBCD only computes the partial gradient of one block coordinate of one mini-batch samples. ORBCD is well suited for the composite minimization problem where one function is the average of the losses of a large number of samples and the other is a simple regularizer defined on high dimensional variables. We show that the iteration complexity of ORBCD has the same order as OGD or SGD. For strongly convex functions, by reducing the variance of stochastic gradients, we show that ORBCD can converge at a geometric rate in expectation, matching the convergence rate of SGD with variance reduction and RBCD. 
\end{abstract}

\section{Introduction}
In recent years, considerable efforts in machine learning have been devoted to solving the following composite objective minimization problem:
\begin{align}\label{eq:compositeobj}
\min_{\x}~f(\x) + g(\x) = \frac{1}{I}\sum_{i=1}^{I}f_i(\x) + \sum_{j=1}^{J}g_j(\x_j)~,
\end{align}
where $\x\in\R^{n\times 1}$ and $\x_j$ is a block coordinate of $\x$. $f(\x)$ is the average of some smooth functions, and $g(\x)$ is a \emph{simple} function which may be non-smooth. In particular, $g(\x)$ is block separable and blocks are  non-overlapping.  A variety of machine learning and statistics problems can be cast into the problem~\myref{eq:compositeobj}. In regularized risk minimization problems~\cite{hastie09:statlearn}, $f$ is the average of losses of a large number of samples and $g$ is a simple regularizer on high dimensional features to induce structural sparsity~\cite{bach11:sparse}. While $f$ is separable among samples, $g$ is separable among features. 
For example, in lasso~\cite{tibs96:lasso}, $f_i$ is a square loss or logistic loss function and $g(\x) = \lambda \| \x \|_1$ where $\lambda$ is the tuning parameter. In group lasso~\cite{yuan07:glasso},  $g_j(\x_j) = \lambda\| \x_j \|_2$, which enforces group sparsity among variables. To induce both group sparsity and sparsity, sparse group lasso~\cite{friedman:sglasso} uses composite regularizers $g_j(\x_j) = \lambda_1\| \x_j \|_2 + \lambda_2 \|\x_j\|_1$ where $\lambda_1$ and $\lambda_2$ are the tuning parameters.

Due to the simplicity, gradient descent (GD) type methods have been widely used to solve problem~\myref{eq:compositeobj}. If $g_j$ is nonsmooth but simple enough for \emph{proximal mapping}, it is better to just use the gradient of $f_i$ but keep $g_j$ untouched in GD. This variant of GD is often called proximal splitting~\cite{comb09:prox} or proximal gradient descent (PGD)~\cite{tseng08:apgm,beck09:pgm} or forward/backward splitting method (FOBOS)~\cite{duchi09}. Without loss of generality, we simply use GD to represent GD and its variants in the rest of this paper. Let $m$ be the number of samples and $n$ be dimension of features. $m$ samples are divided into $I$ blocks (mini-batch), and $n$ features are divided into $J$ non-overlapping blocks.  
If both $m$ and $n$ are large, solving~\myref{eq:compositeobj} using batch methods like gradient descent (GD) type methods is computationally expensive. 
To address the computational bottleneck, two types of low cost-per-iteration methods, online/stochastic gradient descent (OGD/SGD)~\cite{Robi51:SP,Judi09:SP,celu06,Zinkevich03,haak06:logregret,Duchi10_comid,duchi09,xiao10} and randomized block coordinate descent (RBCD)~\cite{nesterov10:rbcd,bkbg11:pbcd,rita13:pbcd,rita12:rbcd}, have been rigorously studied in both theory and applications.

Instead of computing gradients of all samples in GD at each iteration, OGD/SGD only computes the gradient of one block samples, and thus the cost-per-iteration is just 
one $I$-th of GD. For large scale problems, it has been shown that OGD/SGD is faster than GD~\cite{tari13:pdsvm,shsisr07:pegasos,shte09:sgd}.  OGD and SGD have been generalized to handle composite objective functions~\cite{nest07:composite,comb09:prox,tseng08:apgm,beck09:pgm,Duchi10_comid,duchi09,xiao10}. OGD and SGD use a decreasing step size and converge at a slower rate than GD. In stochastic optimization, the slow convergence speed is caused by the variance of stochastic gradients due to random samples, and
considerable efforts have thus been devoted to reducing the variance to accelerate SGD~\cite{bach12:sgdlinear,bach13:sgdaverage,xiao14:psgdvd,zhang13:sgdvd,jin13:sgdmix,jin13:sgdlinear}. 
Stochastic average gradient (SVG)~\cite{bach12:sgdlinear} is the first SGD algorithm achieving the linear convergence rate for stronly convex functions, catching up with the  convergence speed of GD~\cite{nesterov04:convex}. However, SVG needs to store all gradients, which becomes an issue for large scale datasets. It is also difficult to understand the intuition behind the proof of SVG. To address the issue of storage and better explain the faster convergence,~\cite{zhang13:sgdvd} proposed an explicit variance reduction scheme into SGD.  The two scheme SGD is refered as stochastic variance reduction gradient (SVRG). SVRG computes the full gradient periodically and progressively mitigates the variance of stochastic gradient by removing the difference between the full gradient and stochastic gradient. For smooth and strongly convex functions, SVRG converges at a geometric rate in expectation. Compared to SVG, SVRG is free from the storage of full gradients and has a much simpler proof. The similar idea was also proposed independently by~\cite{jin13:sgdmix}. The results of SVRG is then improved in~\cite{kori13:ssgd}. In~\cite{xiao14:psgdvd}, SVRG is generalized to solve composite minimization problem by incorporating the variance reduction technique into proximal gradient method.  

On the other hand, RBCD~\cite{nesterov10:rbcd,rita12:rbcd,luxiao13:rbcd,shte09:sgd,chang08:bcdsvm,hsieh08:dcdsvm,osher09:cdcs} has become increasingly popular due to high dimensional problem with structural regularizers. RBCD randomly chooses a block coordinate to update at each iteration. The iteration complexity of RBCD was established in~\cite{nesterov10:rbcd}, improved and generalized to composite minimization problem by~\cite{rita12:rbcd,luxiao13:rbcd}. RBCD can choose a constant step size and converge at the same rate as GD, although the constant is usually $J$ times worse~\cite{nesterov10:rbcd,rita12:rbcd,luxiao13:rbcd}. Compared to GD, the cost-per-iteration of RBCD is much cheaper.
Block coordinate descent (BCD) methods have also been studied under a deterministic cyclic order~\cite{sate13:cbcd,tseng01:ds,luo02:cbcd}.  Although the convergence of cyclic BCD has been established~\cite{tseng01:ds,luo02:cbcd}, the iteration of complexity is still unknown except for special cases~\cite{sate13:cbcd}.

While OGD/SGD is well suitable for problems with a large number of samples, RBCD is suitable for high dimension problems with non-overlapping composite regularizers. For large scale high dimensional problems with non-overlapping composite regularizers, it is not economic enough to use one of them.  Either method alone may not suitable for problems when data is distributed across space and time or partially available at the moment~\cite{nesterov10:rbcd}. In addition, SVRG is not suitable for problems when the computation of full gradient at one time is expensive.  In this paper,
we propose a new method named online randomized block coordinate descent (ORBCD) which combines the well-known OGD/SGD and RBCD together. ORBCD first randomly picks up one block samples and one block coordinates, then performs the block coordinate gradient descent on the randomly chosen samples at each iteration. Essentially, ORBCD performs RBCD in the online and stochastic setting. 
If $f_i$ is a linear function, the cost-per-iteration of ORBCD is $O(1)$ and thus is far smaller than $O(n)$ in OGD/SGD and $O(m)$ in RBCD. 
We show that the iteration complexity for ORBCD has the same order as OGD/SGD. 
In the stochastic setting, ORBCD is still suffered from the variance of stochastic gradient. To accelerate the convergence speed of ORBCD, we adopt the varaince reduction technique~\cite{zhang13:sgdvd} to alleviate the effect of randomness. 
As expected, the linear convergence rate for ORBCD with variance reduction (ORBCDVD) is established for strongly convex functions for stochastic optimization. Moreover, ORBCDVD does not necessarily require to compute the full gradient at once which is necessary in SVRG and prox-SVRG. Instead,  a block coordinate of full gradient is computed at each iteration and then stored for the next retrieval in ORBCDVD.

The rest of the paper is organized as follows. In Section~\ref{sec:relate}, we review the SGD and RBCD. ORBCD and ORBCD with variance reduction are proposed in Section~\ref{sec:orbcd}. The convergence results are given in Section~\ref{sec:theory}.  The paper is concluded in Section~\ref{sec:conclusion}. 

\section{Related Work}\label{sec:relate}
In this section, we briefly review the two types of low cost-per-iteration gradient descent (GD) methods, i.e., OGD/SGD and RBCD. Applying GD on~\myref{eq:compositeobj}, we have the following iterate:
\begin{align}\label{eq:fobos}
\x^{t+1} = \argmin_{\x}~\langle \nabla f(\x^t), \x \rangle + g(\x) + \frac{\eta_t}{2} \| \x - \x^t \|_2^2~.
\end{align}
In some cases, e.g. $g(\x)$ is $\ell_1$ norm,~\myref{eq:fobos} can have a closed-form solution. 
\subsection{Online and Stochastic Gradient Descent}
In~\myref{eq:fobos}, it requires to compute the full gradient of $m$ samples at each iteration, which could be computationally expensive if $m$ is too large. Instead, OGD/SGD simply computes the gradient of one block samples. 

In the online setting, at time $t+1$, OGD first presents a solution $\x^{t+1}$ by solving 
\begin{align}
\x^{t+1} = \argmin_{\x}~\langle \nabla f_t(\x^t), \x \rangle + g(\x) + \frac{\eta_t}{2} \| \x - \x^t \|_2^2~.
\end{align}
where $f_t$ is given and assumed to be convex. Then a function $f_{t+1}$ is revealed which incurs the loss $f_t(\x^t)$. 
The performance of OGD is measured by the regret bound, which is the discrepancy between the cumulative loss over $T$ rounds and the best decision in hindsight, 
\begin{align}
R(T) = \sum_{t=1}^{T} { [f_t(\x^t) + g(\x^t)] - [f_t(\x^*)+g(\x^*)]}~,
\end{align}
where $\x^*$ is the best result in hindsight. The regret bound of OGD is $O(\sqrt{T})$ when using decreasing step size $\eta_t = O(\frac{1}{\sqrt{t}})$. For strongly convex functions, the regret bound of OGD is $O(\log T)$ when using the step size $\eta_t = O(\frac{1}{t})$. Since $f_t$ can be any convex function, OGD considers the worst case and thus the mentioned regret bounds are optimal.

In the stochastic setting, SGD first randomly picks up $i_t$-th block samples and then computes the gradient of the selected samples as follows:
\begin{align}\label{eq:sgd}
\x^{t+1} = \argmin_{\x}~\langle \nabla f_{i_t}(\x^t), \x \rangle + g(\x) + \frac{\eta_t}{2} \| \x - \x^t \|_2^2~.
\end{align}
$\x^t$ depends on the observed realization of the random variable $\xi = \{ i_1, \cdots, i_{t-1}\}$ or generally $\{ \x^1, \cdots, \x^{t-1} \}$. Due to the effect of variance of stochastic gradient, SGD has to choose decreasing step size, i.e., $\eta_t = O(\frac{1}{\sqrt{t}})$, leading to slow convergence speed. For general convex functions, SGD converges at a rate of $O(\frac{1}{\sqrt{t}})$. For strongly convex functions, SGD converges at a rate of $O(\frac{1}{t})$. In contrast, GD converges linearly if functions are strongly convex. 

To accelerate the SGD by reducing the variance of stochastic gradient,  stochastic variance reduced gradient (SVRG) was proposed by~\cite{zhang13:sgdvd}.~\cite{xiao14:psgdvd} extends SVRG to composite functions~\myref{eq:compositeobj}, called prox-SVRG. SVRGs have two stages, i.e., outer stage and inner stage. The outer stage maintains an estimate $\tilde{\x}$ of the optimal point $x^*$ and computes the full gradient of $\tilde{\x}$
\begin{align} 
\tilde{\mu} &= \frac{1}{n} \sum_{i=1}^{n} \nabla f_i(\tilde{\x}) = \nabla f(\tilde{\x})~.
\end{align}
After the inner stage is completed, the outer stage updates $\tilde{\x}$. At the inner stage, SVRG first randomly picks $i_t$-th sample, then modifies stochastis gradient by subtracting the difference between the full gradient and stochastic gradient at $\tilde{\x}$, 
\begin{align} 
\v_{t} &= \nabla f_{i_t}(\x^t) - \nabla f_{i_t}(\tilde{\x}) + \tilde{\mu}~.
\end{align}
It can be shown that the expectation of $\v_{t}$ given $\x^{t-1}$ is the full gradient at $\x^t$, i.e., $\mathbb{E}\v_{t} = \nabla f(\x^t)$. Although $\v_t$ is also a stochastic gradient, the variance of stochastic gradient progressively decreases. Replacing $\nabla f_{i_t}(\x^t)$ by $\v_t$ in SGD step~\myref{eq:sgd}, 
\begin{align}
\x^{t+1} & = \argmin_{\x}~\langle \v_{t}, \x \rangle + g(\x) + \frac{\eta}{2} \| \x - \x^t \|_2^2~.
\end{align}
By reduding the variance of stochastic gradient, $\x^t$ can converge to $\x^*$ at the same rate as GD, which has been proved in~\cite{zhang13:sgdvd,xiao14:psgdvd}. 
 For strongly convex functions, prox-SVRG~\cite{xiao14:psgdvd} can converge linearly in expectation if $\eta > 4L$ and $m$ satisfy the following condition:
\begin{align}\label{eq:svrg_rho}
\rho = \frac{\eta^2}{\gamma(\eta-4L)m} + \frac{4L(m+1)}{(\eta-4L)m} < 1~.
\end{align}
where $L$ is the constant of Lipschitz continuous gradient. Note the step size is $1/\eta$ here. 

\subsection{Randomized Block Coordinate Descent}
Assume $\x_{j} (1\leq j \leq J)$ are non-overlapping blocks. At iteration $t$, RBCD~\cite{nesterov10:rbcd,rita12:rbcd,luxiao13:rbcd} randomly picks $j_t$-th coordinate  and solves the following problem:
\begin{align}\label{eq:rbcd}
\x_{j_t}^{t+1} = \argmin_{\x_{j_t}}~\langle \nabla_{j_t} f(\x^t), \x_{j_t} \rangle + g_{j_t}(\x_{j_t}) + \frac{\eta_t}{2} \| \x_{j_t} - \x_{j_t}^t \|_2^2~.
\end{align}
Therefore, $\x^{t+1} = (\x_{j_t}^{t+1}, \x_{k\neq j_t}^t)$. $\x^t$ depends on the observed realization of the random variable
\begin{align}
\xi = \{ j_1, \cdots, j_{t-1}\}~.
\end{align} 
Setting the step size $\eta_t  = L_{j_t}$ where $L_{j_t}$ is the Lipshitz constant of $j_t$-th coordinate of the gradient $\nabla f(\x^t)$,  the iteration complexity of RBCD is
 $O(\frac{1}{t})$. For strongly convex function, RBCD has a linear convergence rate. Therefore, RBCD converges at the same rate as GD, although the constant is $J$ times larger~\cite{nesterov10:rbcd,rita12:rbcd,luxiao13:rbcd}.

\section{Online  Randomized Block Coordinate Descent}\label{sec:orbcd}
In this section, our goal is to combine OGD/SGD and RBCD together to solve problem~\myref{eq:compositeobj}.  We call the algorithm online randomized block coordinate descent (ORBCD), which computes one block coordinate of the gradient of one block of samples at each iteration. ORBCD essentially performs RBCD in online and stochastic setting. 

Let $\{ \x_1, \cdots, \x_J \}, \x_j\in \R^{n_j\times 1}$ be J non-overlapping blocks of $\x$. 
Let $U_j \in \R^{n\times n_j}$ be $n_j$ columns of an $n\times n$ permutation matrix $\bU$, corresponding to $j$ block coordinates in $\x$. For any partition of $\x$ and $\bU$,
\begin{align}
\x = \sum_{j=1}^{J}U_j\x_j~, \x_j = U_j^T\x~.
\end{align}
The $j$-th coordinate of gradient of $f$ can be denoted as  
\begin{align}
\nabla_j f(\x) = U_j^T \nabla f(\x)~.
\end{align}
Throughout the paper, we assume that the minimum of problem~\myref{eq:compositeobj} is attained. In addition, ORBCD needs the following assumption : 
\vspace{-3mm}
\begin{asm}\label{asm:orbcd1}
$f_t$ or $f_i$ has block-wise Lipschitz continuous gradient with constant $L_j$, e.g.,
\begin{align}
\| \nabla_j f_t(\x + U_j h_j ) - \nabla_j f_t(\x) \|_2 \leq L_j \| h_j \|_2 \leq L \| h_j \|_2~,
\end{align}
where $L = \max_j L_j$.
\end{asm}
\begin{asm}\label{asm:orbcd2}
1. $\| \nabla f_t (\x^t) \|_2 \leq R_f $, or $\| \nabla f (\x^t) \|_2 \leq R_f $;

2. $\x^t$ is assumed in a bounded set $\cX$, i.e., $\sup_{\x,\y \in \cX} \| \x - \y \|_2 = D$.
\end{asm}
While the Assumption~\ref{asm:orbcd1} is used in RBCD, the Assumption~\ref{asm:orbcd2} is used in OGD/SGD. We may assume the sum of two functions is strongly convex.
\begin{asm}\label{asm:orbcd3}
$f_t(\x) + g(\x)$ or $f(\x) + g(\x)$ is $\gamma$-strongly convex, e.g.,
we have
\begin{align}\label{eq:stronggcov}
f_t(\x) + g(\x) \geq f_t(\y)  + g(\y) + \langle \nabla f_t(\y) + g'(\y), \x - \x^{t} \rangle +  \frac{\gamma}{2}  \| \x - \y \|_2^2~.
\end{align}
where $\gamma > 0$ and $g'(\y)$ denotes the subgradient of $g$ at $\y$.
\end{asm}

\subsection{ORBCD for Online Learning}
In online setting, ORBCD considers the worst case and runs at rounds. 
At time $t$, given any function $f_t$ which may be agnostic, ORBCD randomly chooses $j_t$-th block coordinate and presents the solution by solving the following problem:
\begin{align}\label{eq:orbcdo}
\x_{j_t}^{t+1} &= \argmin_{\x_{j_t}}~\langle \nabla_{j_t} f_t(\x^t), \x_{j_t} \rangle + g_{j_t}(\x_{j_t}) + \frac{\eta_t}{2} \| \x_{j_t} - \x_{j_t}^t \|_2^2 \nonumber \\
& = \text{Prox}_{g_{j_t}}(\x_{j_t} -\frac{1}{\eta_t}\nabla_{j_t} f_t(\x^t) )~,
\end{align}
where $\text{Prox}$ denotes the proximal mapping. If $f_t$ is a linear function, e.g., $f_t = l_t\x^t$, then $\nabla_{j_t} f_t(\x^t) = l_{j_t}$, so solving~\myref{eq:orbcdo} is $J$ times cheaper than OGD. 
Thus, $\x^{t+1} = ( \x_{j_t}^{t+1}, \x_{k\neq j_t}^t)$, or
\begin{align}
\x^{t+1} = \x^t + U_{j_t}(\x_{j_t}^{t+1} - \x_{j_t}^t)~.
\end{align}
Then, ORBCD receives a loss function $f_{t+1}(\x)$ which incurs the loss $f_{t+1}(\x^{t+1})$. The algorithm is summarized in Algorithm~\ref{alg:orbcd_online}. 

$\x^t$ is independent of $j_t$ but depends on the sequence of observed realization of the random variable 
\begin{align}
\xi = \{  j_1, \cdots, j_{t-1} \}.
\end{align}
Let $\x^*$ be the best solution in hindsight. The regret bound of ORBCD is defined as
\begin{align}
R(T) =  \sum_{t=1}^T\left \{ \mathbb{E}_{\xi}[ f_t(\x^t) + g(\x^t) ] - [f_t(\x^*) +g(\x^*)] \right \}~.
\end{align}
By setting $\eta_t = \sqrt{t} + L$ where $L=\max_jL_j$, the regret bound of ORBCD is $O(\sqrt{T})$. For strongly convex functions, the regret bound of ORBCD is $O(\log T)$ by setting $\eta_t = \frac{\gamma t}{J} + L$.

\begin{algorithm*}[tb]
\caption{Online Randomized Block Coordinate Descent for Online Learning}
\label{alg:orbcd_online}
\begin{algorithmic}[1]
\STATE {\bfseries Initialization:} $\x^1 = \mathbf{0}$
\FOR{$t=1 \text{ to } T$}
\STATE randomly pick up $j_t$ block coordinates
\STATE $\x_{j_t}^{t+1} = \argmin_{\x_{j_t} \in \cX_j}~\langle \nabla_{j_t} f_t(\x^t), \x_{j_t} \rangle + g_{j_t}(\x_{j_t}) + \frac{\eta_t}{2} \| \x_{j_t} - \x_{j_t}^t \|_2^2$~
\STATE $\x^{t+1} = \x^t + U_{j_t}(\x_{j_t}^{t+1} - \x_{j_t}^t)$ 
\STATE receives the function $f_{t+1}(\x) + g(\x)$ and incurs the loss $f_{t+1}(\x^{t+1}) + g(\x^{t+1})$
\ENDFOR
\end{algorithmic}
\end{algorithm*}

\subsection{ORBCD for Stochastic Optimization}
In the stochastic setting, ORBCD first randomly picks up $i_t$-th block sample and then randomly chooses $j_t$-th block coordinate. The algorithm has the following iterate:
\begin{align}\label{eq:orbcds}
\x_{j_t}^{t+1} & = \argmin_{\x_{j_t}}~\langle \nabla_{j_t} f_{i_t}(\x^t), \x_{j_t} \rangle + g_{j_t}(\x_{j_t}) + \frac{\eta_t}{2} \| \x_{j_t} - \x_{j_t}^t \|_2^2 \nonumber \\
& = \text{Prox}_{g_{j_t}}(\x_{j_t} -\nabla_{j_t} f_{i_t}(\x^t) )~.
\end{align}
For high dimensional problem with non-overlapping composite regularizers, solving~\myref{eq:orbcds} is computationally cheaper than solving~\myref{eq:sgd} in SGD.
The algorithm of ORBCD in both settings is summarized in Algorithm~\ref{alg:orbcd_stochastic}.

$\x^{t+1}$ depends on $(i_t, j_t)$, but $j_{t}$ and $i_{t}$ are independent.
$\x^t$ is independent of $(i_t, j_t)$ but depends on the observed realization of the random variables 
\begin{align}
\xi = \{ ( i_1, j_1), \cdots, (i_{t-1}, j_{t-1}) \}~.
\end{align}
The online-stochastic conversion rule~\cite{Duchi10_comid,duchi09,xiao10} still holds here. The iteration complexity of ORBCD can be obtained by dividing the regret bounds in the online setting by $T$. Setting $\eta_t = \sqrt{t} + L$ where $L=\max_jL_j$, the iteration complexity of ORBCD is 
\begin{align}
\mathbb{E}_{\xi}  [ f(\bar{\x}^t) + g(\bar{\x}^t) ] - [f(\x) +g(\x)] \leq O(\frac{1}{\sqrt{T}})~.
\end{align}
For strongly convex functions,  setting $\eta_t = \frac{\gamma t}{J} + L$,
\begin{align}
\mathbb{E}_{\xi}  [ f(\bar{\x}^t) + g(\bar{\x}^t) ] - [f(\x) +g(\x)] \leq O(\frac{\log T}{T})~.
\end{align}
The iteration complexity of ORBCD match that of SGD. Simiarlar as SGD, the convergence speed of ORBCD is also slowed down by the variance of stochastic gradient. 

\begin{algorithm*}[tb]
\caption{Online Randomized Block Coordinate Descent for Stochastic Optimization}
\label{alg:orbcd_stochastic}
\begin{algorithmic}[1]
\STATE {\bfseries Initialization:} $\x^1 = \mathbf{0}$
\FOR{$t=1 \text{ to } T$}
\STATE randomly pick up $i_t$ block samples and $j_t$ block coordinates
\STATE $\x_{j_t}^{t+1} = \argmin_{\x_{j_t} \in \cX_j}~\langle \nabla_{j_t} f_{i_t}(\x^t), \x_{j_t} \rangle + g_{j_t}(\x_{j_t}) + \frac{\eta_t}{2} \| \x_{j_t} - \x_{j_t}^t \|_2^2$~ 
\STATE $\x^{t+1} = \x^t + U_{j_t}(\x_{j_t}^{t+1} - \x_{j_t}^t)$ 
\ENDFOR
\end{algorithmic}
\end{algorithm*}

\begin{algorithm*}[tb]
\caption{Online Randomized Block Coordinate Descent with Variance Reduction}
\label{alg:orbcdvd}
\begin{algorithmic}[1]
\STATE {\bfseries Initialization:} $\x^1 = \mathbf{0}$
\FOR{$t=2 \text{ to } T$}
\STATE $\x_0 = \tilde{\x} = \x^t$.
\FOR{$k = 0\textbf{ to } m-1$}
\STATE randomly pick up $i_k$ block samples
\STATE randomly pick up $j_k$ block coordinates
\STATE $\v_{j_k}^{i_k} = \nabla_{j_k} f_{i_k}(\x^{k}) - \nabla_{j_k} f_{i_k}(\tilde{\x}) + \tilde{\mu}_{j_k}$ where $\tilde{\mu}_{j_k} = \nabla_{j_k} f(\tilde{\x})$
\STATE $\x_{j_k}^{k} = \argmin_{\x_{j_k} }~\langle \v_{j_k}^{i_k}, \x_{j_k} \rangle + g_{j_k}(\x_{j_k}) + \frac{\eta_k}{2} \| \x_{j_k} - \x_{j_k}^{k}\|_2^2$~ 
\STATE $\x^{k+1} = \x^k + U_{j_k}(\x_{j_j}^{k+1} - \x_{j_k}^k)$ 
\ENDFOR
\STATE $\x^{t+1} = \x^m$ or $\frac{1}{m}\sum_{k=1}^{m}\x^k$
\ENDFOR
\end{algorithmic}
\end{algorithm*}
\subsection{ORBCD with variance reduction}
In the stochastic setting, we apply the variance reduction technique~\cite{xiao14:psgdvd,zhang13:sgdvd} to accelerate the rate of convergence of ORBCD, abbreviated as ORBCDVD. As SVRG and prox-SVRG, ORBCDVD consists of two stages. At time $t+1$, the outer stage maintains an estimate $\tilde{\x} = \x^t$ of the optimal $\x^*$ and updates $\tilde{\x}$ every $m+1$ iterations. 
The inner stage takes $m$ iterations which is indexed by $k = 0,\cdots, m-1$. At the $k$-th iteration, ORBCDVD randomly picks $i_k$-th sample and  $j_k$-th coordinate and compute
\begin{align}
\v_{j_k}^{i_k} &= \nabla_{j_k} f_{i_k}(\x^k) - \nabla_{j_k} f_{i_k}(\tilde{\x}) + \tilde{\mu}_{j_k}~, \label{eq:orbcdvd_vij} 
\end{align}
where 
\begin{align}\label{eq:orbcdvd_mu} 
\tilde{\mu}_{j_k} = \frac{1}{n} \sum_{i=1}^{n} \nabla_{j_k} f_i(\tilde{\x}) = \nabla_{j_k} f(\tilde{\x})~.
\end{align}
$\v_{j_t}^{i_t}$ depends on $(i_t, j_t)$, and $i_t$ and $j_t$ are independent. Conditioned on $\x^k$, taking expectation over $i_k, j_k$ gives
\begin{align}
\mathbb{E}\v_{j_k}^{i_k} &= \mathbb{E}_{i_k} \mathbb{E}_{j_k}[\nabla_{j_k} f_{i_k}(\x^k) - \nabla_{j_k} f_{i_k}(\tilde{\x}) + \tilde{\mu}_{j_k}] \nonumber \\
&= \frac{1}{J}\mathbb{E}_{i_k}  [\nabla f_{i_k}(\x^k) - \nabla f_{i_k}(\tilde{\x}) + \tilde{\mu} ] \nonumber \\
& = \frac{1}{J}\nabla f(\x^k)~.
\end{align}
Although $\v_{j_k}^{i_k}$ is stochastic gradient, the variance $\mathbb{E} \| \v_{j_k}^{i_k} - \nabla_{j_k} f(\x^k) \|_2^2$ decreases progressively and is smaller than $\mathbb{E} \| \nabla f_{i_t}(\x^t) - \nabla f(\x^t) \|_2^2$. 
Using the variance reduced gradient $\v_{j_k}^{i_k}$, ORBCD then performs RBCD as follows:
\begin{align}
\x_{j_k}^{k+1} &= \argmin_{\x_{j_k}}~ \langle \v_{j_k}^{i_k}, \x_{j_k} \rangle + g_{j_k}(\x_{j_k}) + \frac{\eta}{2} \| \x_{j_k} - \x_{j_k}^k \|_2^2 \label{eq:orbcdvd_xj}~.
\end{align}
After $m$ iterations, the outer stage updates $\x^{t+1}$ which is either $\x^m$ or $\frac{1}{m}\sum_{k=1}^{m}\x^k$. The algorithm is summarized in Algorithm~\ref{alg:orbcdvd}. At the outer stage,
ORBCDVD does not necessarily require to compute the full gradient at once. If the computation of full gradient requires substantial computational
eﬀorts, SVRG has to stop and complete the full gradient step before making progress. In contrast,  $\tilde{\mu}$ can be partially computed at each iteration and then stored for the next retrieval in ORBCDVD.

Assume $\eta > 2L$ and $m$ satisfy the following condition:
\begin{align}\label{eq:orbcdvd_rho}
\rho = \frac{L(m+1)}{(\eta-2L)m} + \frac{(\eta-L)J}{(\eta-2L)m} - \frac{1}{m}+ \frac{\eta (\eta-L)J}{(\eta-2L)m\gamma} < 1~,
\end{align}
Then $h(\x)$ converges linearly in expectation, i.e.,
\begin{align}
 \mathbb{E}_{\xi} [f(\x^t) + g(\x^t) - (f(\x^*) + g(\x^*) ] \leq O(\rho^t)~.
\end{align}

Setting $\eta = 4L$ in~\myref{eq:orbcdvd_rho} yields
\begin{align}
\rho = \frac{m+1}{2m} + \frac{3J}{2m} - \frac{1}{m}+ \frac{6JL}{m\gamma} \leq \frac{1}{2} + \frac{3 J}{2m}(1+\frac{4 L}{\gamma})~.
\end{align}
Setting $m = 18JL/\gamma$, then 
\begin{align}
\rho \leq \frac{1}{2} + \frac{1}{12}(\frac{\gamma}{L}+4) \approx \frac{11}{12}~.
\end{align}
where we assume $\gamma/L \approx 1$ for simplicity.

\section{The Rate of Convergence}\label{sec:theory}
The following lemma is a key building block of the proof of the convergence of ORBCD in both online and stochastic setting.
\begin{lem}
Let the Assumption~\ref{asm:orbcd1} and \ref{asm:orbcd2} hold.
Let $\x^t$ be the sequences generated by ORBCD. $j_t$ is sampled randomly and uniformly from $\{1,\cdots, J \}$. We have
\begin{align}\label{eq:orbcd_key_lem}
& \langle \nabla_{j_t} f_t(\x^t) + g'_{j_t}(\x_{j_t}^t), \x_{j_t}^{t} - \x_{j_t} \rangle   \leq \frac{\eta_t}{2} ( \| \x - \x^t \|_2^2 - \| \x - \x^{t+1} \|_2^2)  +  \frac{R_f^2}{2(\eta_t - L)}   + g(\x^t) - g(\x^{t+1})  ~.
\end{align}
where $L = \max_j L_j$.
\end{lem}
\proof
The optimality condition is
\begin{align}
\langle \nabla_{j_t} f_t(\x^t) + \eta_t (\x_{j_t}^{t+1} - \x_{j_t}^t) + g'_{j_t}(\x_{j_t}^{t+1}),  \x_{j_t}^{t+1} - \x_{j_t} \rangle \leq 0~.
\end{align}
Rearranging the terms yields
\begin{align}
& \langle \nabla_{j_t} f_t(\x^t) + g'_{j_t}(\x_{j_t}^{t+1}) ,  \x_{j_t}^{t+1} - \x_{j_t} \rangle  \leq - \eta_t \langle  \x_{j_t}^{t+1} - \x_{j_t}^t ,  \x_{j_t}^{t+1} - \x_{j_t} \rangle \nonumber \\
& \leq \frac{\eta_t}{2} ( \| \x_{j_t} - \x_{j_t}^t \|_2^2 - \| \x_{j_t} - \x_{j_t}^{t+1} \|_2^2 - \| \x_{j_t}^{t+1} - \x_{j_t}^t \|_2^2 ) \nonumber \\
& = \frac{\eta_t}{2} ( \| \x - \x^t \|_2^2 - \| \x - \x^{t+1} \|_2^2 - \| \x_{j_t}^{t+1} - \x_{j_t}^t \|_2^2 )  ~,
\end{align}
where the last equality uses $\x^{t+1} = (\x_{j_t}^{t+1}, \x_{k\neq {j_t}}^t)$.
By the smoothness of $f_t$, we have
\begin{align}
f_t(\x^{t+1}) \leq f_t(\x^t) + \langle \nabla_j f_t(\x^t), \x_j^{t+1} - \x_j^t \rangle + \frac{L_j}{2} \| \x_j^{t+1} - \x_j^t \|_2^2~.
\end{align}
Since $\x^{t+1}  - \x^t = U_{j_t}(\x_{j_t}^{t+1} - \x_{j_t}^{t})$, 
we have
\begin{align}
& f_t(\x^{t+1}) + g(\x^{t+1}) - [f_t(\x^t) + g(\x^t)] \nonumber \\
& \leq \langle \nabla_{j_t} f_t(\x^t), \x_{j_t}^{t+1} - \x_{j_t}^t \rangle + \frac{L_{j_t}}{2} \| \x_{j_t}^{t+1} - \x_{j_t}^t \|_2^2 + g_{j_t}(\x_{j_t}^{t+1}) - g_{j_t}(\x_{j_t}) + g_{j_t}(\x_{j_t}^{t}) - g_{j_t}(\x_{j_t}) \nonumber \\
& \leq \langle \nabla_{j_t} f_t(\x^t) + g'_{j_t}(\x_{j_t}^{t+1}), \x_{j_t}^{t+1} - \x_{j_t} \rangle +  \frac{L_{j_t}}{2} \| \x_{j_t}^{t+1} - \x_{j_t}^t \|_2^2  - \langle \nabla_{j_t} f_t(\x^t) + g'_{j_t}(\x_{j_t}^{t}), \x_{j_t}^{t} - \x_{j_t} \rangle \nonumber \\
& \leq \frac{\eta_t}{2} ( \| \x - \x^t \|_2^2 - \| \x - \x^{t+1} \|_2^2) +  \frac{L_{j_t} - \eta_t}{2} \| \x_{j_t}^{t+1} - \x_{j_t}^t \|_2^2 - \langle \nabla_{j_t} f_t(\x^t) + g'_{j_t}(\x_{j_t}^{t}), \x_{j_t}^{t} - \x_{j_t} \rangle ~.
\end{align}
Rearranging the terms yields
\begin{align}\label{eq:lem1}
 \langle \nabla_{j_t} f_t(\x^t) + g_{j_t}'(\x^t) , \x_{j_t}^{t} - \x_{j_t} \rangle  &\leq \frac{\eta_t}{2} ( \| \x - \x^t \|_2^2 - \| \x - \x^{t+1} \|_2^2) +  \frac{L_{j_t} - \eta_t}{2} \| \x_{j_t}^{t+1} - \x_{j_t}^t \|_2^2  \nonumber \\
&+  f_t(\x^t)  + g(\x^t)-  [ f_t(\x^{t+1}) + g(\x^{t+1}) ]~.
\end{align}
The convexity of $f_t$ gives
\begin{align}
f_t(\x^t) -  f_t(\x^{t+1})  \leq \langle \nabla f_t(\x^t) , \x^t -  \x^{t+1} \rangle =  \langle \nabla_{j_t} f_t(\x^t) , \x_{j_t}^t -  \x_{j_t}^{t+1} \rangle \leq \frac{1}{2\alpha} \| \nabla_{j_t} f_t(\x^t) \|_2^2 + \frac{\alpha}{2} \| \x_{j_t}^t -  \x_{j_t}^{t+1} \|_2^2~.
\end{align}
where the equality uses $\x^{t+1} = (\x_{j_t}^{t+1}, \x_{k\neq {j_t}}^t)$. Plugging into~\myref{eq:lem1}, we have
\begin{align}
& \langle \nabla_{j_t} f_t(\x^t) + g'_{j_t}(\x_{j_t}^t), \x_{j_t}^{t} - \x_{j_t} \rangle   \nonumber \\
& \leq \frac{\eta_t}{2} ( \| \x - \x^t \|_2^2 - \| \x - \x^{t+1} \|_2^2) +  \frac{L_{j_t} - \eta_t}{2} \| \x_{j_t}^{t+1} - \x_{j_t}^t \|_2^2  +  \langle \nabla_{j_t} f_t(\x^t) , \x_{j_t}^t -  \x_{j_t}^{t+1} \rangle + g(\x^t) - g(\x^{t+1}) \nonumber \\
& \leq \frac{\eta_t}{2} ( \| \x - \x^t \|_2^2 - \| \x - \x^{t+1} \|_2^2) +  \frac{L_{j_t} - \eta_t}{2} \| \x_{j_t}^{t+1} - \x_{j_t}^t \|_2^2  + \frac{\alpha}{2}\| \x_{j_t}^t -  \x_{j_t}^{t+1} \|_2^2   +  \frac{1}{2\alpha} \| \nabla_{j_t} f_t(\x^t) \|_2^2 ~.
\end{align}
Let $ L = \max_{j} L_j$. Setting $\alpha = \eta_t - L$ where $\eta_t > L$ completes the proof.
\qed

This lemma is also a key building block in the proof of iteration complexity of GD, OGD/SGD and RBCD. In GD, by setting $\eta_t = L$, the iteration complexity of GD can be established. In RBCD, by simply setting $\eta_t = L_{j_t}$, the iteration complexity of RBCD can be established.

\subsection{Online Optimization}
Note $\x^t$ depends on the sequence of observed realization of the random variable 
$\xi = \{  j_1, \cdots, j_{t-1} \}$. 
The following theorem establishes the regret bound of ORBCD.
\begin{thm}
Let $\eta_t = \sqrt{t} + L$ in the ORBCD and the Assumption~\ref{asm:orbcd1} and \ref{asm:orbcd2} hold.  $j_t$ is sampled randomly and uniformly from $\{1,\cdots, J \}$. The regret bound $R(T)$ of ORBCD is 
\begin{align}
R(T) \leq  J (  \frac{\sqrt{T} + L}{2}D^2 +  \sqrt{T} R^2 +   g(\x^1) - g(\x^*) )~.
\end{align}
\end{thm}
\proof
In~\myref{eq:orbcd_key_lem}, conditioned on $\x^t$, take expectation over $j_t$, we have
\begin{align}\label{eq:a}
\frac{1}{J}  \langle \nabla f_t(\x^t) + g'(\x^t), \x^{t} - \x \rangle  &\leq \frac{\eta_t}{2} ( \| \x - \x^t \|_2^2 - \mathbb{E}\| \x - \x^{t+1} \|_2^2)  +  \frac{R^2}{2(\eta_t - L)}  + g(\x^t) - \mathbb{E}g(\x^{t+1}) 
~.
\end{align}
Using the convexity, we have
\begin{align}
f_t(\x^t) + g(\x^t) - [f_t(\x) + g(\x)] \leq \langle \nabla f_t(\x^t) + g'(\x^t), \x^{t} - \x \rangle~.
\end{align}
Together with~\myref{eq:a}, we have
\begin{align}
f_t(\x^t) + g(\x^t) - [f_t(\x) + g(\x) ]   &\leq J \left \{ \frac{\eta_t}{2} ( \| \x - \x^t \|_2^2 - \mathbb{E}\| \x - \x^{t+1} \|_2^2) +  \frac{R^2}{2(\eta_t - L)} + g(\x^t) - \mathbb{E}g(\x^{t+1})  \right \}~.
\end{align}
Taking expectation over $\xi$ on both sides, we have
\begin{align}
\mathbb{E}_{\xi} \left [ f_t(\x^t) + g(\x^t) - [f_t(\x) + g(\x) ] \right ]   &\leq J \left \{ \frac{\eta_t}{2} ( \mathbb{E}_{\xi}\| \x - \x^t \|_2^2 - \mathbb{E}_{\xi}\| \x - \x^{t+1} \|_2^2) \right .\nonumber \\
& +  \left. \frac{R^2}{2(\eta_t - L)} + \mathbb{E}_{\xi} g(\x^t) - \mathbb{E}_{\xi}g(\x^{t+1})  \right \}~.
\end{align}
Summing over $t$ and setting $\eta_t = \sqrt{t} + L$, we obtain the regret bound
\begin{align}\label{eq:orbcd_rgt0}
& R(T) = \sum_{t=1}^T\left \{ \mathbb{E}_{\xi}  [ f_t(\x^t) + g(\x^t) ] - [f_t(\x) +g(\x)] \right \} \nonumber \\
&\leq J \left \{ - \frac{\eta_{T}}{2} \mathbb{E}_{\xi}\| \x - \x^{T+1} \|_2^2  + \sum_{t=1}^{T}(\eta_{t} - \eta_{t-1}) \mathbb{E}_{\xi}\| \x - \x^{t} \|_2^2 +  \sum_{t=1}^{T}\frac{R^2}{2(\eta_t - L)}  + g(\x^1) - \mathbb{E}_{\xi}g(\x^{T+1}) \right \} \nonumber \\
& \leq J \left \{  \frac{\eta_T}{2} D^2  +  \sum_{t=1}^{T}\frac{R^2}{2(\eta_t - L)} +   g(\x^1) - g(\x^*) \right \}  \nonumber \\
& \leq J \left \{ \frac{\sqrt{T} + L}{2} D^2  +  \sum_{t=1}^{T}\frac{R^2}{2\sqrt{t} } + g(\x^1) - g(\x^*) \right \} \nonumber \\
& \leq  J (  \frac{\sqrt{T} + L}{2} D^2+  \sqrt{T} R^2 +   g(\x^1) - g(\x^*) )~,
\end{align}
which completes the proof.
\qed

If one of the functions is strongly convex, ORBCD can achieve a $\log(T)$ regret bound, which is established in the following theorem.
\begin{thm}\label{thm:orbcd_rgt_strong}
Let the Assumption~\ref{asm:orbcd1}-\ref{asm:orbcd3} hold and $\eta_t = \frac{\gamma t}{J} + L$ in ORBCD. $j_t$ is sampled randomly and uniformly from $\{1,\cdots, J \}$. The regret bound $R(T)$ of ORBCD is 
\begin{align}
R(T) \leq   J^2R^2 \log(T) +   J(g(\x^1) - g(\x^*) ) ~.
\end{align}
\end{thm}
\proof
Using the strong convexity of $f_t + g$ in~\myref{eq:stronggcov}, we have
\begin{align}
f_t(\x^t) + g(\x^t) - [f_t(\x) + g(\x)] \leq \langle \nabla f_t(\x^t) + g'(\x^t), \x^{t} - \x \rangle - \frac{\gamma}{2} \| \x - \x^t \|_2^2~.
\end{align}
Together with~\myref{eq:a}, we have
\begin{align}
f_t(\x^t) + g(\x^t) - [f_t(\x) + g(\x) ]   &\leq \frac{J\eta_t - \gamma }{2} \| \x - \x^t \|_2^2 - \frac{J\eta_t}{2} \mathbb{E}\| \x - \x^{t+1} \|_2^2) \nonumber \\
& +  \frac{JR^2}{2(\eta_t - L)} + J [ g(\x^t) - \mathbb{E}g(\x^{t+1}) ] ~.
\end{align}
Taking expectation over $\xi$ on both sides, we have
\begin{align}
\mathbb{E}_{\xi} \left [ f_t(\x^t) + g(\x^t) - [f_t(\x) + g(\x) ] \right ]   &\leq \frac{J\eta_t - \gamma}{2} \mathbb{E}_{\xi}\| \x - \x^t \|_2^2 - \frac{J\eta_t}{2}\mathbb{E}_{\xi}[\| \x - \x^{t+1} \|_2^2]) \nonumber \\
& +   \frac{JR^2}{2(\eta_t - L)} + J [ \mathbb{E}_{\xi} g(\x^t) - \mathbb{E}_{\xi}g(\x^{t+1})  ]~.
\end{align}
Summing over $t$ and setting $\eta_t = \frac{\gamma t}{J} + L$,  we obtain the regret bound
\begin{align}
& R(T) = \sum_{t=1}^T\left \{ \mathbb{E}_{\xi}  [ f_t(\x^t) + g(\x^t) ] - [f_t(\x) +g(\x)] \right \} \nonumber \\
&\leq   - \frac{J\eta_{T}}{2} \mathbb{E}_{\xi}\| \x - \x^{T+1} \|_2^2  + \sum_{t=1}^{T}\frac{J\eta_{t} -\gamma - J\eta_{t-1}}{2}\mathbb{E}_{\xi}\| \x - \x^{t} \|_2^2 +  \sum_{t=1}^{T}\frac{JR^2}{2(\eta_t - L)}  + J ( g(\x^1) - \mathbb{E}_{\xi}g(\x^{T+1}) ) \nonumber \\
& \leq \sum_{t=1}^{T}\frac{J^2R^2}{2\gamma t } +   J(g(\x^1) - g(\x^*) )  \nonumber \\
& \leq J^2R^2 \log(T) +   J(g(\x^1) - g(\x^*) )  ~,
\end{align}
which completes the proof.
\qed

In general, ORBCD can achieve the same order of regret bound as OGD and other first-order online optimization methods, although the constant could be $J$ times larger.

\subsection{Stochastic Optimization}
In the stochastic setting, ORBCD first randomly chooses the $i_t$-th block sample and the $j_t$-th block coordinate.
$j_{t}$ and $i_{t}$ are independent. $\x^t$ depends on the observed realization of the random variables
$\xi = \{ ( i_1, j_1), \cdots, (i_{t-1}, j_{t-1}) \}$.
The following theorem establishes the iteration complexity of ORBCD for general convex functions.
\begin{thm}\label{thm:orbcd_stc_ic}
Let $\eta_t = \sqrt{t} + L$ and $\bar{\x}^T = \frac{1}{T} \sum_{t=1}^{T}\x^t $ in the ORBCD. $i_t, j_t$ are sampled randomly and uniformly from $\{1,\cdots, I \}$ and $\{1,\cdots, J \}$ respectively. The iteration complexity of ORBCD is 
\begin{align}
\mathbb{E}_{\xi}  [ f(\bar{\x}^t) + g(\bar{\x}^t) ] - [f(\x) +g(\x)] \leq  \frac{J (  \frac{\sqrt{T} + L}{2} D^2+  \sqrt{T} R^2 +   g(\x^1) - g(\x^*) )}{T}~.
\end{align}
\end{thm}
\proof
In the stochastic setting, let $f_t$ be $f_{i_t}$ in~\myref{eq:orbcd_key_lem}, we have
\begin{align}\label{eq:orbcd_key_stoc}
\langle \nabla_{j_t} f_{i_t}(\x^t) + g_{j_t}'(\x^t) , \x_{j_t}^{t} - \x_{j_t} \rangle  \leq \frac{\eta_t}{2} ( \| \x - \x^t \|_2^2 - \| \x - \x^{t+1} \|_2^2)  +  \frac{R^2}{2(\eta_t - L)}   + g(\x^t) - g(\x^{t+1})  ~.
\end{align}
Note $i_t, j_t$ are independent of $\x^t$. Conditioned on $\x^t$, taking expectation over $i_t$ and $j_t$, the RHS is
\begin{align}
& \mathbb{E}\langle \nabla_{j_t} f_{i_t}(\x^t) + g_{j_t}'(\x^t) , \x_{j_t}^{t} - \x_{j_t} \rangle  = \mathbb{E}_{i_t} [ \mathbb{E}_{j_t} [ \langle \nabla_{j_t} f_{i_t}(\x^t) + g_{j_t}'(\x^t), \x_{j_t}^{t} - \x_{j_t} \rangle ]  ] \nonumber \\
& = \frac{1}{J}  \mathbb{E}_{i_t} [ \langle  \nabla f_{i_t}(\x^t), \x^{t} - \x \rangle + \langle g'(\x^t) , \x^{t} - \x \rangle ] \nonumber \\
& = \frac{1}{J}\langle  \nabla f(\x^t) + g'(\x^t) , \x^{t} - \x \rangle ~.
\end{align}
Plugging back into~\myref{eq:orbcd_key_stoc}, we have
\begin{align}\label{eq:orbcd_stc_0}
& \frac{1}{J}\langle  \nabla f(\x^t) + g'(\x^t) , \x^{t} - \x \rangle \nonumber \\
&\leq \frac{\eta_t}{2} ( \| \x - \x^t \|_2^2 - \mathbb{E}\| \x - \x^{t+1} \|_2^2) +  \frac{R^2}{2(\eta_t - L)}   + g(\x^t) - \mathbb{E} g(\x^{t+1}) ~.
\end{align}
Using the convexity of $f + g$, we have
\begin{align}
f(\x^t) + g(\x^t) - [f(\x) + g(\x)] \leq \langle \nabla f(\x^t) + g'(\x^t), \x^{t} - \x \rangle~.
\end{align}
Together with~\myref{eq:orbcd_stc_0}, we have
\begin{align}
f(\x^t) + g(\x^t) - [f(\x) + g(\x) ]   &\leq J \left \{ \frac{\eta_t}{2} ( \| \x - \x^t \|_2^2 - \mathbb{E}\| \x - \x^{t+1} \|_2^2) +  \frac{R^2}{2(\eta_t - L)} + g(\x^t) - \mathbb{E}g(\x^{t+1})  \right \}~.
\end{align}
Taking expectation over $\xi$ on both sides, we have
\begin{align}
\mathbb{E}_{\xi} \left [ f(\x^t) + g(\x^t) \right ] - [f(\x) + g(\x) ]   &\leq J \left \{ \frac{\eta_t}{2} ( \mathbb{E}_{\xi}\| \x - \x^t \|_2^2 - \mathbb{E}_{\xi}[\| \x - \x^{t+1} \|_2^2]) \right .\nonumber \\
& +  \left. \frac{R^2}{2(\eta_t - L)} + \mathbb{E}_{\xi} g(\x^t) - \mathbb{E}_{\xi}g(\x^{t+1})  \right \}~.
\end{align}
Summing over $t$ and setting $\eta_t = \sqrt{t} + L$, following similar derivation in~\myref{eq:orbcd_rgt0}, we have
\begin{align}
\sum_{t=1}^T\left \{ \mathbb{E}_{\xi}  [ f(\x^t) + g(\x^t) ] - [f(\x) +g(\x)] \right \} \leq  J (  \frac{\sqrt{T} + L}{2} D^2+  \sqrt{T} R^2 +   g(\x^1) - g(\x^*) )~.
\end{align}
Dividing both sides by $T$, using the Jensen's inequality and denoting $\bar{\x}^T = \frac{1}{T}\sum_{t=1}^{T}\x^t$ complete the proof.
\qed

For strongly convex functions, we have the following results.
\begin{thm}
For strongly convex function, setting $\eta_t = \frac{\gamma t}{J} + L$ in the ORBCD. $i_t, j_t$ are sampled randomly and uniformly from $\{1,\cdots, I \}$ and $\{1,\cdots, J \}$ respectively. Let $\bar{\x}^T = \frac{1}{T} \sum_{t=1}^{T}\x^t $. The iteration complexity of ORBCD is 
\begin{align}
\mathbb{E}_{\xi}  [ f(\bar{\x}^T) + g(\bar{\x}^T) ] - [f(\x) +g(\x)] \leq   \frac{J^2R^2 \log(T) +   J(g(\x^1) - g(\x^*) ) }{T}~.
\end{align}
\end{thm}
\proof
If $f+g$ is strongly convex, we have
\begin{align}
f(\x^t) + g(\x^t) - [f(\x) + g(\x)] \leq \langle \nabla f(\x^t) + g'(\x^t), \x^{t} - \x \rangle - \frac{\gamma}{2} \| \x - \x^t \|_2^2~.
\end{align}
Plugging back into~\myref{eq:orbcd_stc_0}, following similar derivation in Theorem~\ref{thm:orbcd_rgt_strong} and Theorem~\ref{thm:orbcd_stc_ic} complete the proof.
\qed

\subsection{ORBCD with Variance Reduction}
According to the Theorem 2.1.5 in~\cite{nesterov04:convex}, the block-wise Lipschitz gradient in Assumption~\ref{asm:orbcd1} can also be rewritten as follows:
\begin{align}
& f_i(\x)  \leq f_i(\y) + \langle \nabla_j f_i(\x) - \nabla_j f_i(\y), \x_j - \y_j\rangle + \frac{L}{2} \| \x_j - \y_j\|_2^2~, \label{eq:blk_lip1} \\
&\| \nabla_j f_i(\x) - \nabla_j f_i(\y) \|_2^2 \leq L \langle \nabla_j f_i(\x) - \nabla_j f_i(\y), \x_j - \y_j\rangle~.\label{eq:blk_lip2}
\end{align}

Let $\x^*$ be an optimal solution. Define an upper bound of $f(\x) + g(\x) -(f(\x^*)+g(\x^*))$ as 
\begin{align}\label{eq:def_h}
h(\x,\x^*)= \langle \nabla f(\x), \x - \x^*\rangle + g(\x) - g(\x^*)~.
\end{align}
 If $f(\x) + g(\x)$ is strongly convex, we have
\begin{align}\label{eq:orbcd_strong_h}
h(\x,\x^*) \geq f(\x) - f(\x^*) + g(\x) - g(\x^*) \geq \frac{\gamma}{2} \| \x - \x^* \|_2^2 ~.
\end{align}

\begin{lem}\label{lem:orbcdvd_lem1}
Let $\x^*$ be an optimal solution and the Assumption~\ref{asm:orbcd1}, we have
\begin{align}
\frac{1}{I} \sum_{i=1}^{I} \| \nabla f_i(\x) - \nabla f_i(\x^*) \|_2^2 \leq L h(\x,\x^*)~.
\end{align}
where $h$ is defined in~\myref{eq:def_h}.
\end{lem}
\proof Since the Assumption~\ref{asm:orbcd1} hold, we have using 
\begin{align}\label{eq:orbcd_bd_h}
&\frac{1}{I} \sum_{i=1}^{I} \| \nabla f_i(\x) - \nabla f_i(\x^*) \|_2^2 = \frac{1}{I} \sum_{i=1}^{I} \sum_{j=1}^J \| \nabla_j f_i(\x) - \nabla_j f_i(\x^*) \|_2^2 \nonumber \\
&\leq \frac{1}{I} \sum_{i=1}^{I} \sum_{j=1}^J L \langle \nabla_j f_i(\x) - \nabla_j f_i(\x^*), \x_j - \x_j^*\rangle \nonumber \\
& = L [ \langle \nabla f(\x), \x - \x^*\rangle + \langle \nabla f(\x^*),  \x^* - \x \rangle]~,
\end{align}
where the inequality uses~\myref{eq:blk_lip2}. For an optimal solution $\x^*$, $g'(\x^*) + \nabla f(\x^*) = 0$ where $g'(\x^*)$ is the subgradient of $g$ at $\x^*$.  The second term in~\myref{eq:orbcd_bd_h} can be rewritten as
\begin{align}
& \langle \nabla f(\x^*),  \x^* - \x \rangle = - \langle g'(\x^*),  \x^* - \x \rangle  = g(\x) - g(\x^*) ~.
\end{align}
Plugging into~\myref{eq:orbcd_bd_h} and using~\myref{eq:def_h} complete the proof.
\qed

\begin{lem}\label{lem:orbcdvd_lem2}
Let $\v_{j_k}^{i_k} $ and $\x_{j_k}^{k+1}$ be generated by~\myref{eq:orbcdvd_vij}-\myref{eq:orbcdvd_xj}. Conditioned on $\x^k$, we have
\begin{align}
\mathbb{E} \| \v_{j_k}^{i_k} - \nabla_{j_k} f(\x^k) \|_2^2  \leq \frac{2L}{J} [h(\x^{k},\x^*) + h(\tilde{\x},\x^*)]~. 
\end{align}
\end{lem}
\proof
Conditioned on $\x^k$, we have
\begin{align}
\mathbb{E}_{i_k}[ \nabla f_{i_k}(\x^k) - \nabla f_{i_k}(\tilde{\x}) + \tilde{\mu} ] = \frac{1}{I} \sum_{i=1}^{I} [ \nabla f_{i}(\x^k) - \nabla f_{i}(\tilde{\x}) + \tilde{\mu} ] = \nabla f(\x^k)~.
\end{align}
Note $\x^k$ is independent of $i_k, j_k$. $i_k$ and  $j_k$ are independent. Conditioned on $\x^k$, taking expectation over $i_k, j_k$ and using~\myref{eq:orbcdvd_vij} give
\begin{align}
&\mathbb{E}\| \v_{j_k}^{i_k} - \nabla_{j_k} f(\x^k) \|_2^2 = \mathbb{E}_{i_k} [ \mathbb{E}_{j_k}\| \v_{j_k}^{i_k} - \nabla_{j_k} f(\x^k) \|_2^2] \nonumber \\
&= \mathbb{E}_{i_k}[ \mathbb{E}_{j_k}\| \nabla_{j_k} f_{i_k}(\x^k) - \nabla_{j_k} f_{i_k}(\tilde{\x}) + \tilde{\mu}_{j_k} - \nabla_{j_k} f(\x^k) \|_2^2] \nonumber \\
&= \frac{1}{J}\mathbb{E}_{i_k}\| \nabla f_{i_k}(\x^k) - \nabla f_{i_k}(\tilde{\x}) + \tilde{\mu} - \nabla f(\x^k) \|_2^2 \nonumber \\
& \leq  \frac{1}{J} \mathbb{E}_{i_k}\| \nabla f_{i_k}(\x^k) - \nabla f_{i_k}(\tilde{\x}) \|_2^2 \nonumber \\
& \leq \frac{2}{J} \mathbb{E}_{i_k}\| \nabla f_{i_k}(\x^k) - \nabla f_{i_k}(\x^*) \|_2^2 + \frac{2}{J} \mathbb{E}_{i_k}\| \nabla f_{i_k}(\tilde{\x}) - \nabla f_{i_k}(\x^*) \|_2^2 \nonumber \\
& = \frac{2}{IJ} \sum_{i=1}^{I}\| \nabla f_i(\x^k) - \nabla f_i(\x^*) \|_2^2 + \frac{2}{IJ}\sum_{i=1}^{I} \| \nabla f_i(\tilde{\x}) - \nabla f_i(\x^*) \|_2^2 \nonumber \\
& \leq \frac{2L}{J} [ h(\x^{k}, \x^*) + h(\tilde{\x}, \x^*)]~.
\end{align}
The first inequality uses the fact $\mathbb{E} \| \zeta - \mathbb{E}\zeta  \|_2^2 \leq \mathbb{E} \| \zeta \|_2^2$ given a random variable $\zeta$, the second inequality uses $\| \a + \b \|_2^2 \leq 2 \| \a \|_2^2 + 2\|\b\|_2^2$, and the last inequality uses Lemma~\ref{lem:orbcdvd_lem1}.
\qed

\begin{lem}\label{lem:orbcdvd_lem3}
Under Assumption~\ref{asm:orbcd1}, $f(\x) = \frac{1}{I} \sum_{i=1}^{I}f_i(\x)$ has block-wise Lipschitz continuous gradient with constant $L$, i.e.,
\begin{align}
\| \nabla_j f(\x + U_j h_j ) - \nabla_j f(\x) \|_2 \leq L \| h_j \|_2~. 
\end{align}
\end{lem}
\proof
Using the fact that $f(\x) = \frac{1}{I} \sum_{i=1}^{I}f_i(\x)$, we have
\begin{align}
&\| \nabla_j f(\x + U_j h_j ) - \nabla_j f(\x) \|_2  = \| \frac{1}{I} \sum_{i=1}^{I} [\nabla_j f_i(\x + U_j h_j ) - \nabla_j f_i(\x) ] \|_2 \nonumber \\
& \leq  \frac{1}{I} \sum_{i=1}^{I} \| \nabla_j f_i(\x + U_j h_j ) - \nabla_j f_i(\x) \|_2 \nonumber \\
& \leq L  \| h_j \|_2~,
\end{align}
where the first inequality uses the Jensen's inequality and the second inequality uses the Assumption~\ref{asm:orbcd1}.
\qed

Now, we are ready to establish the linear convergence rate of ORBCD with variance reduction for strongly convex functions.
\begin{thm}\label{thm:orbcdvd}
Let $\x^t$ be generated by ORBCD with variance reduction~\myref{eq:orbcdvd_mu}-\myref{eq:orbcdvd_xj}. $j_k$ is sampled randomly and uniformly from $\{1,\cdots, J \}$. Assume $\eta > 2L$ and $m$ satisfy the following condition:
\begin{align}
\rho = \frac{L(m+1)}{(\eta-2L)m} + \frac{(\eta-L)J}{(\eta-2L)m} - \frac{1}{m}+ \frac{\eta (\eta-L)J}{(\eta-2L)m\gamma} < 1~,
\end{align}
Then ORBCDVD converges linearly in expectation, i.e.,
\begin{align}
 \mathbb{E}_{\xi} [ f(\x^t) + g(\x^t) - (f(\x^*)+g(\x^*) ]  \leq \rho^t [ \mathbb{E}_{\xi} h(\x^1, \x^*)]~.
\end{align}
where $h$ is defined in~\myref{eq:def_h}.
\end{thm}
\proof
The optimality condition of~\myref{eq:orbcdvd_xj} is
\begin{align}
\langle \v_{j_k}^{i_k}  + \eta (\x_{j_k}^{k+1} - \x_{j_k}^k) + g'_{j_k}(\x_{j_k}^{k+1}),  \x_{j_k}^{k+1} - \x_{j_k} \rangle \leq 0~.
\end{align}
Rearranging the terms yields
\begin{align}
& \langle \v_{j_k}^{i_k}  + g'_{j_k}(\x_{j_k}^{k+1}) ,  \x_{j_k}^{k+1} - \x_{j_k} \rangle  \leq - \eta \langle  \x_{j_k}^{k+1} - \x_{j_k}^k ,  \x_{j_k}^{k+1} - \x_{j_k} \rangle \nonumber \\
& \leq \frac{\eta}{2} ( \| \x_{j_k} - \x_{j_k}^k \|_2^2 - \| \x_{j_k} - \x_{j_k}^{k+1} \|_2^2 - \| \x_{j_k}^{k+1} - \x_{j_k}^k \|_2^2 ) \nonumber \\
& = \frac{\eta}{2} ( \| \x - \x^k \|_2^2 - \| \x - \x^{k+1} \|_2^2 - \| \x_{j_k}^{k+1} - \x_{j_k}^k \|_2^2 )  ~,
\end{align}
where the last equality uses $\x^{k+1} = (\x_{j_k}^{k+1}, \x_{k\neq {j_k}}^t)$.
Using the convecxity of $g_j$ and the fact that $g(\x^k) - g(\x^{k+1}) =  g_{j_k}(\x^k) -  g_{j_k}(\x^{k+1})$, we have
\begin{align}
& \langle \v_{j_k}^{i_k}  ,  \x_{j_k}^{k} - \x_{j_k} \rangle  + g_{j_k}(\x^k) - g_{j_k}(\x)  \leq \langle \v_{j_k}^{i_k} ,  \x_{j_k}^{k} - \x_{j_k}^{k+1} \rangle + g(\x^k) - g(\x^{k+1}) \nonumber \\
& + \frac{\eta}{2} ( \| \x - \x^k \|_2^2 - \| \x - \x^{k+1} \|_2^2 - \| \x_{j_k}^{k+1} - \x_{j_k}^k \|_2^2 ) ~.
\end{align}
According to Lemma~\ref{lem:orbcdvd_lem3} and using~\myref{eq:blk_lip1}, we have
\begin{align}
\langle \nabla_{j_k} f(\x^k), \x_{j_k}^{k} - \x_{j_k}^{k+1} \rangle \leq f(\x^k) - f(\x^{k+1}) + \frac{L}{2} \| \x_{j_k}^{k} - \x_{j_k}^{k+1} \|_2^2~.
\end{align}
Letting $\x = \x^*$ and using the smoothness of $f$, we have
\begin{align}
& \langle \v_{j_k}^{i_k}  ,  \x_{j_k}^{k} - \x_{j_k} \rangle  + g_{j_k}(\x^k) - g_{j_k}(\x^*)  \leq \langle \v_{j_k}^{i_k} - \nabla_{j_k} f(\x^k),  \x_{j_k}^{k} - \x_{j_k}^{k+1} \rangle +  f(\x^k) + g(\x^k) - [f(\x^{k+1})+g(\x^{k+1})] \nonumber \\
& + \frac{\eta}{2} ( \| \x^* - \x^k \|_2^2 - \| \x^* - \x^{k+1} \|_2^2 - \| \x_{j_k}^{k+1} - \x_{j_k}^k \|_2^2 )  + \frac{L}{2} \| \x_{j_k}^{k} - \x_{j_k}^{k+1} \|_2^2\nonumber \\
& \leq \frac{1}{2(\eta-L)} \| \v_{j_k}^{i_k} - \nabla_{j_k} f(\x^k) \|_2^2  +  f(\x^k) + g(\x^k) - [f(\x^{k+1})+g(\x^{k+1})] + \frac{\eta}{2} ( \| \x^* - \x^k \|_2^2 - \| \x^* - \x^{k+1} \|_2^2 ) ~.
\end{align}
Taking expectation over $i_k, j_k$ on both sides and using Lemma~\ref{lem:orbcdvd_lem2}, we have
\begin{align}\label{eq:orbcdvd_expbd}
& \mathbb{E} [ \langle \v_{j_k}^{i_k}  ,  \x_{j_k}^{k} - \x_{j_k}^* \rangle  + g_{j_k}(\x^k) - g_{j_k}(\x^*)]  \nonumber \\
&\leq \frac{L}{J(\eta-L)} [h(\x^k,\x^*) + h(\tilde{\x},\x^*)] + f(\x^k) + g(\x^k) - \mathbb{E}[f(\x^{k+1})+g(\x^{k+1})] \nonumber \\
& + \frac{\eta}{2} ( \| \x^* - \x^k \|_2^2 - \mathbb{E}\| \x^* - \x^{k+1} \|_2^2 )~.
\end{align}
The left hand side can be rewritten as
\begin{align}
& \mathbb{E} [\langle \v_{j_k}^{i_k}  ,  \x_{j_k}^{k} - \x_{j_k}^* \rangle  + g_{j_k}(\x^k) - g_{j_k}(\x^*)]  = \frac{1}{J} [ \mathbb{E}_{i_k}\langle \v^{i_k}  ,  \x^k - \x^* \rangle + g(\x^k) -g(\x^*) ] \nonumber \\
& = \frac{1}{J} [ \langle \nabla f(\x^k) ,  \x^k - \x^* \rangle + g(\x^k) -g(\x^*) ] = \frac{1}{J} h(\x^k,\x^*)~.
\end{align}
Plugging into~\myref{eq:orbcdvd_expbd} gives
\begin{align}
\frac{1}{J} [ h(\x^k,\x^*) ] & \leq \frac{L}{J(\eta-L)} [h(\x^k,\x^*) + h(\tilde{\x},\x^*)] + f(\x^k) + g(\x^k) - \mathbb{E}[f(\x^{k+1})+g(\x^{k+1})]  \nonumber \\
&+ \frac{\eta}{2} ( \| \x^* - \x^k \|_2^2 - \mathbb{E}\| \x^* - \x^{k+1} \|_2^2 ) \nonumber \\
& \leq \frac{L}{J(\eta-L)} [h(\x^k,\x^*) + h(\tilde{\x},\x^*)] + f(\x^k) + g(\x^k) - \mathbb{E}[f(\x^{k+1})+g(\x^{k+1})]  \nonumber \\
&+ \frac{\eta}{2} ( \| \x^* - \x^k \|_2^2 - \mathbb{E}\| \x^* - \x^{k+1} \|_2^2 ) ~,
\end{align}
Rearranging the terms yields
\begin{align}
\frac{\eta - 2L}{J(\eta-L)}  h(\x^k,\x^*)  &\leq \frac{L}{J(\eta-L)}[ h(\tilde{\x},\x^*) ] + f(\x^k) + g(\x^k) - \mathbb{E}[f(\x^{k+1})+g(\x^{k+1})]  \nonumber \\
& + \frac{\eta}{2} ( \| \x^* - \x^k \|_2^2 - \mathbb{E}\| \x^* - \x^{k+1} \|_2^2 )~.
\end{align}
At time $t+1$,  we have $\x_0 = \tilde{\x} = \x^t$.  Summing over $k = 0,\cdots, m$ and taking expectation with respect to the history of random variable $\xi$,  we have
\begin{align}
\frac{\eta - 2L}{J(\eta-L)} \sum_{k=0}^{m} \mathbb{E}_{\xi}h(\x_k,\x^*) &\leq \frac{L(m+1)}{J(\eta-L)} \mathbb{E}_{\xi}h(\tilde{\x},\x^*) + \mathbb{E}_{\xi}[ f(\x_0) + g(\x_0) ] - \mathbb{E}_{\xi} [ f(\x_{m+1}) + g(\x_{m+1})] \nonumber \\
&+ \frac{\eta}{2} ( \mathbb{E}_{\xi}\| \x^* - \x_0 \|_2^2 - \mathbb{E}_{\xi}\| \x^* - \x_{m+1} \|_2^2 ) \nonumber \\
&\leq \frac{Lm}{J(\eta-L)} \mathbb{E}_{\xi}h(\tilde{\x},\x^*) + \mathbb{E}_{\xi}h(\x_0,\x^*) + \frac{\eta}{2} \mathbb{E}_{\xi}\| \x^* - \x_0 \|_2^2 \nonumber ~,
\end{align}
where the last inequality uses 
\begin{align}
f(\x_0) + g(\x_0)  -  [ f(\x_{m+1}) + g(\x_{m+1})] & \leq  f(\x_0) + g(\x_0)  -  [ f(\x^*) + g(\x^*)]  \nonumber \\
& \leq \langle\nabla f(\x_0), \x_0 - \x^* \rangle + g(\x_0) - g(\x^*) \nonumber \\
& = h(\x_0,\x^*)~.
\end{align}
Rearranging the terms gives
\begin{align}
 \frac{\eta -2L}{J(\eta-L)} \sum_{k=1}^{m} \mathbb{E}_{\xi}h(\x^k,\x^*)  \leq \frac{L(m+1)}{J(\eta-L)} \mathbb{E}_{\xi}h(\tilde{\x},\x^*) + (1- \frac{\eta -2 L}{J(\eta-L)} ) \mathbb{E}_{\xi}h(\x_0,\x^*) + \frac{\eta}{2} \mathbb{E}_{\xi}\| \x^* - \x_0 \|_2^2~.
\end{align}
Pick $x^{t+1}$ so that $h(\x^{t+1}) \leq h(\x_k),  1\leq k \leq m$, we have
\begin{align}\label{eq:orbcdvd_lineareq0}
\frac{\eta - 2L}{J(\eta-L)} m \mathbb{E}_{\xi} h(\x^{t+1},\x^*)  \leq [ \frac{L(m+1)}{J(\eta-L)}  + 1- \frac{\eta -2 L}{J(\eta-L)} ] \mathbb{E}_{\xi}h(\x^t,\x^*) + \frac{\eta}{2} \mathbb{E}_{\xi}\| \x^* - \x^t \|_2^2~,
\end{align}
where ther right hand side uses $\x^t = \x_0 = \tilde{\x}$. Using~\myref{eq:orbcd_strong_h}, we have
\begin{align}
\frac{\eta - 2L}{J(\eta-L)} m \mathbb{E}_{\xi} h(\x^{t+1} ,\x^*)  \leq [ \frac{L(m+1)}{J(\eta-L)}  + 1- \frac{\eta -2 L}{J(\eta-L)} +\frac{\eta}{\gamma}  ] \mathbb{E}_{\xi}h(\x^t,\x^*)~.
\end{align}
Dividing both sides by $\frac{\eta - 2L}{J(\eta-L)} m$, we have
\begin{align}
 \mathbb{E}_{\xi} h(\x^{t+1},\x^*)  \leq \rho \mathbb{E}_{\xi}h(\x^t,\x^*)~,
\end{align}
where 
\begin{align}
\rho = \frac{L(m+1)}{(\eta-2L)m} + \frac{(\eta-L)J}{(\eta-2L)m} - \frac{1}{m}+ \frac{\eta (\eta-L)J}{(\eta-2L)m\gamma} < 1~,
\end{align}
which completes the proof.
\qed


\section{Conclusions}\label{sec:conclusion}
We proposed online randomized block coordinate descent (ORBCD) which combines online/stochastic gradient descent and randomized block coordinate descent.  ORBCD is well suitable for large scale high dimensional problems with non-overlapping composite regularizers. We established the rate of convergence for ORBCD, which has the same order as OGD/SGD. For stochastic optimization with strongly convex functions, ORBCD can converge at a geometric rate in expectation by reducing the variance of stochastic gradient. 

\section*{Acknowledgment}
H.W. and A.B. acknowledge the support of NSF via IIS-0953274, IIS-1029711, IIS- 0916750, IIS-0812183, NASA grant NNX12AQ39A, and the technical support from the University of Minnesota Supercomputing Institute. A.B. acknowledges support from IBM and Yahoo. H.W. acknowledges the support of DDF (2013-2014) from the University of Minnesota. H.W. also thanks Renqiang Min and Mehrdad Mahdavi for mentioning the papers about variance reduction when the author was in the NEC Research Lab, America.

\bibliographystyle{plain}
\bibliography{long,bcd,admm,onlinelearn,sparse,map}

\end{document}